\documentclass[10pt,journal,comssoc]{IEEEtran}

\newtheorem{definition}{\textbf{Definition}}
\newtheorem{problem}{\textbf{Problem}}

\newcommand{\figref}[1]{Fig~\ref{fig:#1}}

\newcommand{\tblref}[1]{Table~\ref{tbl:#1}}

\newcommand{\secref}[1]{Section~\ref{sec:#1}}

\newcommand{\alggref}[1]{Algorithm~\ref{alg:#1}}

\newcommand{\defref}[1]{Definition~\ref{def:#1}}

\newcommand{\probref}[1]{Problem~\ref{problem:#1}}

\ifCLASSOPTIONcompsoc
  \usepackage[nocompress]{cite}
\else
  \usepackage{cite}
\fi

\ifCLASSINFOpdf
\else
\fi
\usepackage{graphicx}

\usepackage{amsmath}

\usepackage{algorithm}
\usepackage[noend]{algpseudocode}

\begin{document}
\title{Proximity-Based Active Learning on Streaming Data: A Personalized Eating Moment Recognition}

\author{Marjan~Nourollahi,~
        Seyed~Ali~Rokni,~
        and~Hassan~Ghasemzadeh
\thanks{$^{*}$Marjan Nourollahi, S.A. Rokni, and H. Ghasemzadeh are with the School of Electrical Engineering and Computer Science, Washington State University, Pullman, WA 99164, USA (phone: 509-335-8260; e-mail: \{m.nourollahidarabad, s.roknidezfooli, hassan.ghasemzadeh\}@wsu.edu).}
}

\IEEEtitleabstractindextext{
\begin{abstract}
Detecting when eating occurs is an essential step toward automatic dietary monitoring, medication adherence assessment, and diet-related health interventions. Wearable technologies play a central role in designing unubtrusive diet monitoring solutions by leveraging machine learning algorithms that work on time-series sensor data to detect eating moments. While much research has been done on developing activity recognition and eating moment detection algorithms, the performance of the detection algorithms drops substantially when the model trained with one user is utilized by a new user. To facilitate development of personalized models, we propose PALS\footnote{Software code for PALS is available online at https://github.com/marjan-nourollahi/PALS}, Proximity-based Active Learning on Streaming data, a novel proximity-based model for recognizing eating gestures with the goal of significantly decreasing the need for labeled data with new users. Particularly, we propose an optimization problem to perform active learning under limited query budget by leveraging unlabeled data. Our extensive analysis on data collected in both controlled and uncontrolled settings indicates that the F-score of PLAS ranges from 22\% to 39\% for a budget that varies from 10 to 60 query. Furthermore, compared to the state-of-the-art approaches, off-line PALS, on average, achieves to 40\% higher recall and 12\% higher f-score in detecting eating gestures.
\end{abstract}

}
\maketitle

\IEEEdisplaynontitleabstractindextext
\IEEEpeerreviewmaketitle

\section{Introduction}\label{sec:introduction}
\IEEEPARstart{E}{ating} habits are highly correlated with human health and wellbeing \cite{flegal2002prevalence}. It is not only what people eat that contributes to their health but is also when and how often the eating events occur \cite{nicklas2001eating}. An automatic health monitoring system can help with monitoring eating habits. These systems can also accommodate users with special health conditions such as diabetes \cite{helal2009smart}, those at need to take their medication at certain times during the day such as after or in between a meal, or assist users who need to follow a special dietary plan \cite{chatterjee2009healthy}. Detecting when eating happens is a key challenge in automatic health monitoring. 

Most current approaches for eating moment recognition require multiple on-body sensors or specialized devices \cite{bedri2017earbit,rahman2014bodybeat,bedri2015detecting}, which make these solutions impractical for everyday living scenarios. The aim of this research is to design a machine learning model that uses easy-to-wear and prevalent devices such as smartwatches for eating moment detection.

However, we recognize that different people perform the same activity differently as a result of which relying on a model trained by collected data of one or few subjects will not provide desired accuracy when used with new subjects. A major challenge with customization of the machine learning algorithms is that retraining the model needs large amounts of labeled training data. Collecting enough labeled data is a time consuming, labor-intensive, and expensive process. Considering this fact that user's pattern in performing activities are different in real-life scenarios compared to in-lab settings, the problem becomes even more challenging. A potential approach to collect ground truth labels in real-life scenarios is to continuously record user's activities using a body-worn cameras However, deploying cameras in uncontrolled settings impose serious privacy concerns. Therefore, it is critical to develop strategies that allow for collecting ground truth labels outside laboratory settings.

Active learning is potentially a feasible approach to query sensor data for ground truth labels in end-user settings. Such an approach will allow us to query a small subset of sensor data based on an informativeness measurement \cite{settles2012active} an yet achieve an acceptable accuracy level. However, in mobile health and streaming data situations, the sensors are sampled in real-time and a decision needs to be made instantaneously about querying or skipping a data segment needs to be made. This is an area of research that has remained unexplored by the community. To address the problem of activity learning with streaming sensor data, we propose PALS as a proximity-based active learning approach for eating moment recognition. To the best of our knowledge, PALS is the first attempt to develop a practical approach for eating moment detection using an active learning framework for human-in-the-loop learning on streaming sensor data.

\section{Related Work}
Our work in this article spans two areas of research including (1) diet monitoring; and (2) active learning. In this section, we discuss the state-of-the-art research in each area.

\subsection{Diet Monitoring}
The pervasive nature of new technologies such as smartwatches and light-weight wearable devices with embedded inertial sensors (e.g., accelerometer and gyroscope) has resulted in development of eating moment detection algorithms.
One example of such researches is the work by Thomaz et al. in which authors introduced an eating episodes detection approach using accelerometer data of an off-the-shelf smartwatch \cite{thomaz2015practical}. 
In another study, Tauhidur et al. presented BodyBeat, a custom-built microphone designed to detect non-speech body sounds such as food intake by capturing skin vibration \cite{rahman2014bodybeat}. In another study, bedri et al. explored the use of inertial, optical, and acoustic sensing modalities for eating moment detection \cite{bedri2017earbit}. Furthermore, Dong et al. utilized watch-like configuration of sensors to track hand motion \cite{dong2014detecting}. Proposing the idea that eating episodes tend to be preceded and succeeded by the events of vigorous hand movements, authors used signal energy for classification of the activities \cite{dong2014detecting}. In another study, Yatani et al. presented BodyScope, a wearable acoustic-sensor-based system that uses neck-worn sensor data for diet monitoring \cite{yatani2012bodyscope}. Cheng et al. also explored the use of a neckband to recognize different eating activities \cite{cheng2013activity}.

\subsection{Active Learning}
A challenging task in diet monitoring is to collect sufficient amounts of labeled data in uncontrolled environments for algorithm training. One approach to collect labeled data is to use active learning technologies that query the user to label sensor data in real-time. In general, active learning has shown promising results in achieving a higher accuracy level using less labeled instances \cite{settles2012active}. Active learning has been studied in two major scenarios including pool-based and stream-based cases \cite{settles2012active}. In the pool-based scenario \cite{lewis1994sequential}, a big pool of unlabeled examples are given and an oracle can provide truth label for instances in this pool. A major challenge in stream-based active learning is that the learner does not have access to the future instances. Therefore, the learner needs to decide about the informativeness of the instances in real-time and in absence of forthcoming data. Therefore, in the stream-based  scenario \cite{cohn1994improving}, upon receiving  a new instance, the learner decides whether to query for truth label and update the classifier or ignore the current instance. 

While a fixed uncertainty sampling method has been used in the past to label instances within a batch of data from the data stream \cite{zhu2007active}, {\v{Z}}liobait{\.e} et al. designed a dynamic allocation strategy of labeling with a randomized search space without considering batches \cite{vzliobaite2011active}. In addition to utilizing an evolving model \cite{smailovic2014stream}, ensemble classifiers could be used to decide about the informativeness of instances \cite{zhu2007active,wang2012mining,zhu2010active} by training a number of classifiers on different portions of data stream. While many of these approaches have been proposed to address a concept drift in highly dynamic environments such as Twitter, our approach considers the personalization of the model for its current user in real-time running on a resource limited device such as a smartphone or smartwatch.

Nonetheless, the utility of active learning in diet monitoring with wearable sensors in general, and in scenarios with streaming data in particular, has not been investigated to date. We introduce a proximity-based active learning approach to improve the performance of the model with less labeled data while leveraging unlabeled data for model training. Inspired by graph-based semi-supervised learning research \cite{zhu2006semi,zhu2003semi}, our approach utilizes unlabeled data to improve the quality of the model.

\section{Problem Statement}\label{sec:ps}
Let $\mathcal{X}$ denote a large set of collected sensor data. An observation $\bf{x}_i$ made by a wearable sensor at time $i$ can be represented as a $D$-dimensional feature vector, $\bf{x}_i$ = \{$w_{i1}$, $w_{i2}$, $\dots$, $w_{iD}$\}. Each feature is computed from a given time window and a marginal probability distribution over all possible feature values. The activity recognition task is composed of a label space $\mathcal{A}$=\{$a_1$, $a_2$, $\dots$, $a_m$\} consisting of the set of labels for activities of interest, and a conditional probability distribution $P(\mathcal{A}|\bf{x_i})$ which is the probability of assigning a label $a_j \in \mathcal{A}$ given an observed instance $\bf{x}_i$. Subsequently, the final predicted label for observation $\bf{x}_i$ is defined as 
\begin{equation}
f(\bf{x}_i)= arg\max_{a_j \in \mathcal{A}} P(a_j |\bf{x_i}) 
\end{equation} 

Although, given the growing ubiquity of Internet-of-Things (IoT) sensors, collecting a large pool of unlabeled sensor data is attainable, labeling such a huge amount of data using human supervision is time-consuming, burdensome, and expensive. Therefore, it is important to devise an efficient approach for selecting informative instances taking into account the constraint of limited budget to query an expert for ground truth labels. Furthermore, because the sensors are sampled continuously as the user performs various daily activities, the active learning algorithm needs to select sensor data for query in real-time. The reason for such a constraint is that expecting the user/expert to provide true labels for activities that occurred in the past is subject to human memory and bias errors. Therefore, it is desirable to decide if a query needs to be issued for the currently occurring activity. In this section, we formally define active learning as an optimization problem.

\subsection{Limited Budget Training}

To approach the problem of active learning given both budget and real-time decision making constraints, we first relax the second constraint by assuming that a human expert can label a pool of sensor data collected in the past by either remembering the activities or watching a video recording of the activities. This allows us to develop a basic pool-based active learning algorithm that selects most informative instances from a large pool of the collected sensor data. In the next step, we show how the pool-based algorithm can be modified for realizing real-time active learning scenarios where a decision about querying the expert is made instantaneously. In the following, we formulate each of the problems and present our solution to solve those problems. \probref{LBT} formally defines the limited budget active learning problem.

\begin{problem}[Limited Budget Training (LBT)]\label{problem:LBT}
Assume an active learning algorithm splits the instances in $\mathcal{X}$ into two disjoint subsets $l$ and $U$ where the instances in $l$ are used to query the oracle to obtain their true labels and those in $U$ remain unlabeled. The Limited Budget Training (LBT) problem is to efficiently construct the small subset $l$ and train a classifier such that the error of classifying instances in $\mathcal{X}$ is minimized and the size of $l$ is bounded by a given query budget of $\Delta$. 
\end{problem}

The LBT problem described in \probref{LBT} can be formulated as follows.

\begin{equation} \label{equ:OF}
Minimize \; \; \; \; \; \sum_{i= 1}^{|\mathcal{X}|}(|f(\bf{x_{i}})-y_{i}|)
\end{equation}
\begin{equation} \label{equ:con}
|l| \leq \Delta
\end{equation}

\begin{equation}\label{equ:cup}
l \cup U = \mathcal{X}
\end{equation}
 
\begin{equation} \label{equ:cap}
l \cap U = \emptyset
\end{equation}

The objective function in \eqref{equ:OF} aims to minimize the amount of misclassification error given the budget constraint in \eqref{equ:con}. The constraints in \eqref{equ:cup} and \eqref{equ:cap} are based on the definition where $l$ and $U$ are considered a perfect partitioning of set $\mathcal{X}$.

As described in \probref{LBT}, due to limited budget constraint, designing an efficient method to cherry pick instances to feed the training process is essential. Here, \defref{is} formally defines the instance selector function.

\begin{definition}[Instance Selector] \label{def:is}
An instance selector  $\mathcal{I}$ is a function $\mathcal{I} : \mathcal{X} \rightarrow \{0,1\}$ such that

\begin{equation}
\mathcal{I}=  
\begin{cases}
      1, & \text{if}\ \bf{x_i} \in l\\
      0, & \text{otherwise}
\end{cases}
\end{equation}
\end{definition}

\noindent where $x_i$ refers to the instances selected for query. Considering that the active learning algorithm uses the instance selector $\mathcal{I}$, the \probref{LBT} could be re-formulate as an Integer Linear Programming problem as follows.

\begin{equation} \label{equ:LOF}
Minimize   \; \; \; \; \; \sum_{i=1}^{\left | \mathcal{X} \right |}(1-\mathcal{I}(i)) |f(\bf{x_{i}})-y_{i}|
\end{equation}

\begin{equation} \label{equ:LCON}
\sum_{i=1}^{\left |\mathcal{X} \right |} \mathcal{I}(i) = \Delta
\end{equation}

The objective function in \eqref{equ:LOF} aims to minimize the amount of misclassification error on unknown instances while \eqref{equ:LCON} states the budget constraint.

A major limitation of the LBT problem described above is that it assumes a perfect memory retention for the oracle. That is, the oracle is able to remember the past events reliably. In reality, however, mobile health technologies monitor end users continuously and the user may not remember past events. Therefore, is it more realistic to design an active learning approach for streaming sensor data. In the following, we reformulate \probref{LBT} taking into account that the oracle provides labels for current activity. \probref{LBTS} formally defines the problem of training with limited budget on a stream of data. 

\begin{problem}[Limited Budget Training on Data Stream
 (LBTS)]\label{problem:LBTS}
 
Let $\mathcal{X}$=[$x_1$, $x_2$, $\dots$, $x_t$, $\dots$, $x_T$] be a sequence of sensor instances that are being produced during time frame $t$= \{$1$, $\dots$, $T$\}. An active learning algorithm on stream splits the instances in $\mathcal{X}$ into two disjoint subsequences $l$ and $U$ where the instances in $l$ are used in order to query the oracle to obtain their true label and update the model as they become available in real-time while $U$ remain unlabeled. The Limited Budget Training on Stream(LBTS) is to efficiently decide whether to query the true label for the instance at time $t$ and update the classifier as it becomes available in real-time such that the error of classifying instances in $U$ is minimized.
\end{problem}

Using Linear Programming framework in \eqref{equ:LOF}, the problem of limited budget training on data stream could be formulated as follows.
\begin{equation} \label{equ:SLOF}
Minimize   \; \; \; \; \; \sum_{t=1}^{T}(1-\mathcal{I}(t)) |f_t(\bf{x_{t}})-y_{t}|
\end{equation}

\begin{equation} \label{equ:SLCON}
\sum_{t=1}^{T} \mathcal{I}(t) = \Delta
\end{equation}
\noindent where $f_t$ is the classification function at time $t$. The objective function in \eqref{equ:SLOF} aims to minimize the amount of misclassification error given the budget constraint in \eqref{equ:SLCON}.

\section{PALS Framework Design}
PALS framework focuses on two characteristics of everyday living situations: (1) the ubiquity of data and the ability of obtaining huge amounts of unlabeled data with mobile devices and wearable sensors; and (2) realistic assumption that the user/expert has a limited capability or interest in providing ground truth labels for the massive amounts of data that are being collected in continuous health monitoring applications. Therefore, the general goal of the PALS framework is to leverage the unlabeled data to construct an efficient model while choosing a small subset of instances of the unlabeled data to query the user/expert for label/annotation. In the following, we described our approach for leveraging unlabeled data through a proximity graph model and selecting informative data instances in preparation to query the expert.

\subsection{Proximity-Based Modeling} \label{sec:prox-model}
Inspired by graph-based semi-supervised learning research, we propose to construct a proximity-based model to quantify similarity among data instances. The intuition behind a proximity-based modeling and label inference is \textit{smoothness assumption}. The \textit{smoothness assumption} suggests that the instances that are close in the feature space should have similar labels \cite{zhu2006semi}. The process of constructing a proximity-based model includes two phases. The first phase aims to build a proximity graph using both labeled and unlabeled data. Leveraging unlabeled data could potentially improve the model. As suggested by prior research \cite{zhu2006semi}, in absence of sufficient labeled data, using both labeled and unlabeled data can lead to a more accurate decision boundary for the learned model. The second phase is label inference, which focuses on generating labels for unlabeled instances through an iterative label propagation method.

\begin{definition}[Proximity Graph]
A \textit{proximity graph} $~$ $G(V,E)$ is a weighted graph where each node in $V$ represents an instance in $X = l \cup U$. Each node in the graph maintains a vector of its own feature values and the probability distribution of its labels. An edge $e_{ij} \in E$ represents the amount of similarity between instances $\bf{x_i}$ and $\bf{x_j}$.
\end{definition}



 We denote the similarity between $\bf{x_i}$ and $\bf{x_j}$ by $\eta(\bf{x_i}, \bf{x_j})$ and compute its value by their euclidean distance:
\begin{equation}
\eta(i,j) = \lVert \bf{x_i}- \bf{x_j} \rVert
\end{equation} 

To avoid the confusion of far away instances, we build similarity graph using $k$-NN schema which is one of the most popular approaches in similarity graph construction \cite{k-near}. Therefore, we measure edge weights in the similarity graph using the following equation:

\begin{equation}\label{graph_edge}
e_{ij} = \Big\{
  \begin{tabular}{ll}
  $\eta(i,j)$ &if $i \in \kappa(j)$ or $j \in \kappa(i)$ \\
  $0$  &otherwise
  \end{tabular}
  \Big.
\end{equation}

\noindent where $\kappa(i)$ is the set of $k$-nearest-neighbors of instance $\bf{x_i}$ based on the defined similarity function. 

In practice, we will show that using the $k$-NN schema improves the performance of the trained model in detecting eating moments. 

\subsection{Instance Selector}
To maximize the labeling accuracy while taking into account the constraint in \eqref{equ:con}, we need an effective instance selector function to select the most informative instances from $\mathcal{X}$ to add to the training data used to learn a final model. To quantify informativeness of the instances, in this article, we use an entropy-based method, which generate a score for a given instance based on Information Gain ($IG$) from that instance. Recall that entropy indicates certainty of the model in classifying an instance. An entropy of zero means pure certainty with one of the classes receiving a probability of one. Therefore, low values of entropy suggests that the model is confident about how to classify the input instance. The instance selector $\mathcal{I}$ sorts the instances by their information gain and selects the instance with highest information gain to add to the labeled pool $l$.

\subsection{Off-line PALS}
As described previously in \secref{ps}, in the off-line version of PALS, we assume that a pool of unlabeled sensor instances are available to the oracle. The oracle is then able to label any of instances and to assign the correct activity label upon request. In this off-line approach, we assume that the provided label is correct. This assumption is based on the fact that either the oracle's memory is perfect that they can remember the past events or there is a video recording of the activities that the oracle can navigate to find the correct label for a queried activity.

\begin{figure*}[!tbh]
\centering
  \includegraphics[width=0.75\textwidth]{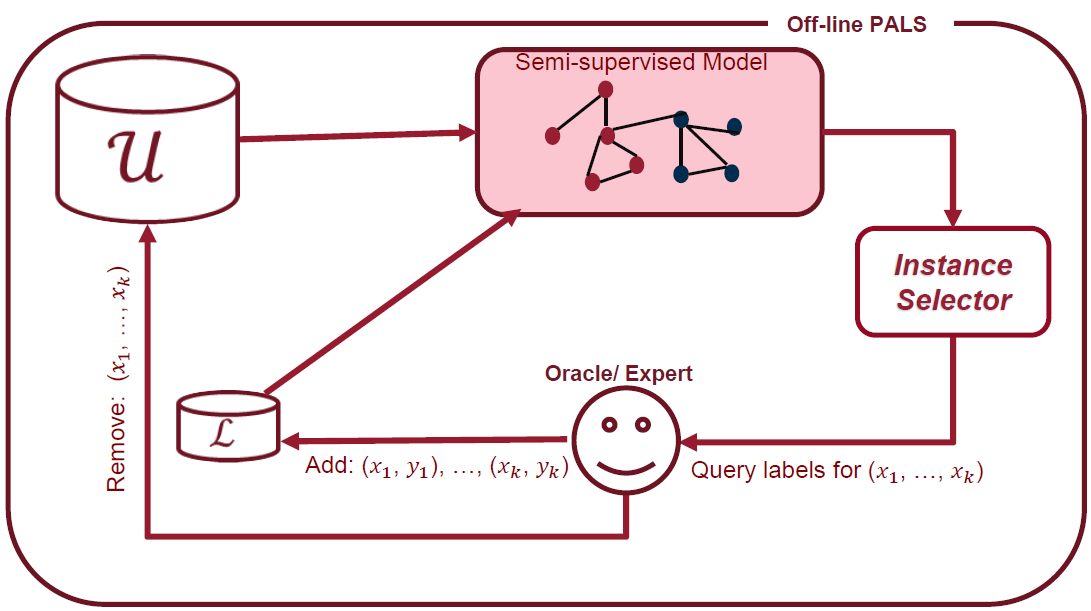}
  \caption{Overall architecture of PALS for off-line active learning.}
  \label{fig:palfig}
\end{figure*}

\figref{palfig} shows the overall architecture of our off-line proximity-based active learning approach. Initially, among all of the recorded activities there is no or a small set of labeled instances $l$ along with a large pool of unlabeled instances $U$. Our algorithm constructs a proximity-based graph on the entire dataset using both $l$ and $U$. Following the graph construction phase, the model aims to infer the actual label of the instances in $U$ in multiple iterations of the label propagation procedure. In the next step, the instance selector $\mathcal{I}$ searches through the unlabeled instances to find the most informative instance in $U$, to date, to request for a label. The process concludes by adding the labeled instance to the model.

\begin{algorithm}
	\caption{Algorithm for Off-line PALS}\label{alg:pal}
	
	\begin{algorithmic}[1]
		\Statex \textbf{Input}: 
		labeled data $l$, 
		unlabeled pool $U$,
		number of iterations $k$, 
		budget $\Delta$
		\Statex \textbf{Output}: Proximity-based model $f$
		\Statex \textbf{Initialize}: $\delta \gets\  \frac {\Delta} {k}$ ,
		$itr \gets k$
		\Procedure {Offline PALS}{} 
		\State $f \ \gets$ construct proximity-based model on $l \cup U$  
		\While{$itr >0$}
		\State $L_u \ \gets$ inferred labels on $U$ using model $f$ 
		\State $IGs \ \gets IG(L_u)$
		\State $X_{sel} \ \gets \delta$ instances with highest $IGs$
		\State $l_{sel} \ \gets$ labels provided by oracle for $X_{sel}$
		\State $l \ \gets l \cup (X_{sel}, l_{sel})$
		\State $f \ \gets$ update model $f$ with new instances in $l$
		\State $itr \gets itr - 1 $
		\EndWhile
		\EndProcedure
	\end{algorithmic}
\end{algorithm}

As illustrated in \figref{palfig}, the process continues iteratively  by obtaining new labeled instances and adding them to the labeled set $l$. The model is then updated and the process of label inference and instance selection are repeated. The algorithm finishes when all the allowed queries are exhausted (i.e. $|l| = \Delta$). \alggref{pal} shows the off-line active learning approach in PALS.

\subsection{Real-time PALS}
To realize real-time active learning on streaming data, we develop real-time PALS. Development of real-time PALS is motivated by the fact that both non-stop video recording of user's activities in naturalistic settings and assuming perfect memory for the user to accurately remember all activities performed in a given time-frame in the past are unrealistic for activity recognition in free living situations. Therefore, to develop a personalized model in a real-life scenarios, we cannot sovely rely on pool-based active learning. Yet, we develop our real-time PALS algorithms based on the foundations established in our off-line PALS.

The main challenge in real-time active learning is to be able to make a decision about whether or not to query each sensor instance as it becomes available in real-time. In particular, because the model does not have access to future instances, it needs to determine whether the current instance is informative enough for which to request a label. Our general approach to make such a determination in real-time is to define a threshold on informativeness of a given instance. Such a threshold, if defined appropriately, will allow us to make real-time active learning decisions.

\begin{definition} [Informativeness Threshold] Let $\mathcal{X}^{sort}$ be the entire stream $\mathcal{X}$ sorted in  informativeness score given by $IG$. An informativeness threshold $\lambda$ is a value such that $IG(\mathcal{X}^{sorted}_{\Delta}) = \lambda$ where $\Delta$ is the query budget.
\end{definition}

\begin{figure*}[!tbh]
	\centering  
	\includegraphics[width=0.75\textwidth]{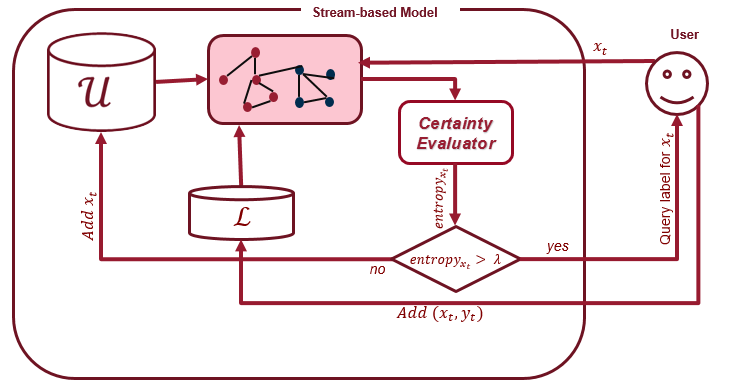}
	\caption{Overall architecture of PALS for real-time active learning on streaming sensor data.}
	\label{fig:spalfig}
\end{figure*}

As shown in \figref{spalfig}, real-time PALS assumes that the user can provide labels only for the current or very recent activities. In this approach, each instance is evaluated only once. As a result of this evaluation, the instance is either discarded from further analysis or used to query the oracle. If the system receives a label from the oracle, the next step is to update the model with the new instance in an effort to obtain a more personalized model. This is accomplished by adding the newly labeled instance to the labeled pool.

\begin{algorithm}
\caption{Algorithm for real-time PALS.}\label{alg:stpal}
\begin{algorithmic}[1]
\Statex \textbf{Input}: current model $f_c$, new instance $x$, threshold $\lambda$, budget $\Delta$
\Statex \textbf{Output}: $f$ 
\Procedure {Real-time PALS}{} 
\State $p\ \gets$ make a prediction on $x$ using model $f_c$  
\State $e\ \gets $ calculate entropy of $p$ 
\If{ $e\ \geq\lambda$ and $\Delta>0$}
\State $\Delta\ \gets \Delta - 1$
\State $y\ \gets$ query oracle to provide true label for $x$
\State $f \gets$ update model $f_c$ with $(x\ ,\ y)$
\EndIf
\EndProcedure
\end{algorithmic}
\end{algorithm}


We need an algorithm to adjust the value of the informativeness threshold to balance labeling over the instance space. In order to obtain an effective performance, the algorithm needs to avoid both high and low values of $\lambda$. High values of $\lambda$ will translate into a highly conservative approach where the a very small number of questions are asked. Therefore, the algorithm can fail in personalizing the model for the current user due to lack of sufficient input from the user. On the other hand, low values of $\lambda$ will result in the algorithm exhausting the budget very quickly rather than generating queries that are distributed in time. Therefore, we need an adaptive algorithm to adjust the value of $\lambda$ to create a balance between prompting time and query budget.


\subsection{Adaptive Threshold Setting}\label{sec:ATS}
An adaptive algorithm for adjusting $\lambda$ needs to address concerns of when and how to update $\lambda$ to achieve an effective performance. Our strategy is to update $\lambda$ after receiving a new instance to a value that ensures a uniform distribution of queries over a given time interval. Suppose $N$ denotes the number of instances over a given time interval. Also assume that we have seen $k$ instances so far. To uniformly distribute queries over the time interval, we need to adjust $\lambda$ taking into account the fact that $\frac {k}{N}$ percentage of the budget has been already exhausted. Here we describe how $\lambda$ can be adjusted for a stream of data to ensure a uniform distribution of queries over a given time interval of $T$.

Let $T$ denote a given time interval over which the active learning process is expected to execute. Also, let $\mathcal{X}_{t}$ represent the data stream generated up to time $t$ and $\mathcal{X}_{t}^{sort}$ be $\mathcal{X}_{t}$ sorted in non-decreasing order by informativeness score given by $IG$. Furthermore, let $\Delta_{t}$ be $(t/T) \times \Delta$. An informativeness threshold at time $t$ is denoted by $\lambda_{t}$ is a value such that $IG((\mathcal{X}_{t})^{sorted}_{\Delta_{t}}) = \lambda_{t}$. The threshold value  $\lambda_{t}$ aims to ensure a uniform distribution of queries over the time interval $T$. This process for obtaining an adaptive $\lambda$ is shown in \alggref{ala}.

\begin{algorithm}[tbh!]
\caption{Algorithm for adaptive adjustment of informativeness threshold, $\lambda$}\label{alg:ala}
\begin{algorithmic}[1]
\Statex \textbf{Input}: current model $m$, new instance $x$, time $t$, time interval $T$, Entropy of instances up to current time $E$, budget $\Delta$
\Statex \textbf{Output}: $\lambda_{t}$ 
\Procedure {Adaptive$\lambda$}{} 
\State $p\ \gets$ use model $m$ to make predictions about $x$ 
\State $e\ \gets $ calculate entropy of $p$ 
\State $E \gets {(E+e)}^{sorted}$
\State $index \gets (t/T) \times \Delta $ 
\State $\lambda_{t} \gets E_{index}$
\EndProcedure
\end{algorithmic}
\end{algorithm}

\section{Validation Approach}
Our goal is to evaluate the performance of PALS using data collected from real subjects performing different activities in both semi-controlled lab settings and free-living environments. In this section, we discuss the datasets, data pre-processing, and performance metrics used for validation of our active learning algorithms. 

\subsection{Data Collection}\label{sec:dataDesc} 
We designed an experiment to collect wearable motion sensor data during eating sessions. The data collection took place between February to April 2017.Institutional Review Board (IRB) approval was obtained prior to data collection. Overall, the dataset contains $20$ sessions of eating data collected with four participants. Each data collection session took about 20 minutes and participants were continuously video recorded. In each session, the participant was asked to eat a meal while performing other related activities such as drinking, talking, working with laptop, and texting. In data collection sessions, participants were asked to wear a Samsung smartwatch on their dominant hand. The smartwach used in our experiment was equipped with a 3D accelerometer and a 3D gyroscope sensors. We developed an android application to sample inertial sensors at $50$ Hz. About $32$\% of the obtained dataset includes eating-related activities. The recorded data for each participant along with labels are  available\footnote{https://github.com/marjan-nourollahi/PALS/dataset} for public use. In the rest of this paper, we refer to this dataset as {\it SW6S} referring to smartwatch dataset with 6 axes of inertial sensor data collected in semi-controlled settings.

\subsection{Publicly Available Dataset}\label{sec:DColl}
To assess the generalization of our approach, we also evaluate the performance of PALS on two publicly available datasets \cite{thomaz2015practical}. Both datasets contain 3D accelerometer data collected from a wirst-band worn on the dominent hand.  The first dataset was collected in a semi-controlled lab setting with $20$ participants performing different activities including eating, watching a movie trailer, chatting, taking a walk, placing a phone call, brushing teeth, and combing hair. In the rest of this paper, we refer to this dataset as {\it SW3S} indicating the smartwatch dataset with 3-axis sensor data collected in semi-controlled settings. The second dataset was collected in free-living settings with seven participants. The participants in this study wore the wrist-band for an average of $5$ hours and $42$ minutes while performing various daily activities such as taking, commuting,  reading, walking, working with a computer, and eating. This dataset includes approximately 6.7\% of eating activity \cite{thomaz2015practical}. In the rest of this paper, we refer to this dataset as {\it SW3U} indicating smartwatch data with 3-axis sensor data in uncontrolled settings.

\subsection{Data Processing Pipeline}
In this section we explain the data processing pipeline and challenges associated with utilizing the sensor data for algorithm development in the context of our active learning research. 

Our data processing pipeline consists of four phases including pre-processing, segmentation, feature extraction, and feature selection. In the pre-processing phase, we pass the raw signal through a low-pass filter to reduce the instrumental noise that generates high-frequency components in the signal. 

The next phase is segmentation which is intended to identify `start' and `end' points of the activity being examined for classification. During the segmentation phase, we use a sliding window with $50$\% overlap to split the continuous signal into segments. The window size is an important parameter because it needs to be long enough to capture an entire food intake gesture. According to previous research, a window length of 6 seconds with $50$\% overlap is a proper segmentation strategy for eating recognition task \cite{thomaz2015practical}.

The next step in the pipeline is feature extraction where we extract $15$ features from each signal segment for each axis of sensor data(e.g., 45 features for SW3S and SW3U and 90 features for SW6S). Potentially, there are many different features that can be extracted from human activity signals. However, as  shown in \tblref{features}, our extracted features can capture both morphology and statistical attributes of the the signals. For example, while features such as {\it median} and {\it mean} capture intensity of the signal, {\it variance} and {\it zero crossing} intend to capture morphology of the signal.

To maximize generalizability of the model and reduce the risk of overfitting, we need to control the complexity of the hypothesis. To this end, we perform {\it feature selection} to identify the best set of features. Particularly, we use $\chi^2$ feature selection method. Similar to statistics domain where $\chi^2$ test is used to test independence of two events, we use this test in our feature selection process to determine whether a specific feature and occurrence of a particular activity are independent. This feature selection approach eliminates irrelevant features from the file feature set.

\begin{table}[tbh!]
\centering
\caption{Features extracted from each signal segment.}
\begin{tabular}{l l} \hline \hline
Feature Name & Description\\\hline \hline
median  & Median value\\ 
mean  & Mean value \\ 
max  & Maximum value\\ 
min  & Minimum value\\ 
p2p  & Peak-to-peak amplitude\\ 
skew  &  Skewness of signal segment \\ 
kurtosis  &  Kurtosis of signal segment \\ 
variance  & Variance of signal segment\\ 
peaks count  & Number of peaks\\ 
mean peaks amplitude   & Mean of peaks amplitude\\ 
max peaks amplitude  & Maximum peak amplitude\\ 
mean peaks distance  & Mean of peaks distance\\ 
min peaks distance  & Minimum of peaks distance\\ 
std peaks distance  & Standard deviation of peaks distance\\ 
zero crossings  & Number of zero crossings\\ \hline \hline
\end{tabular}
\label{tbl:features}
\end{table}

\subsection{Learning from Skewed Data}
In real daily life settings, the duration of activities are not equal among all daily activities. This leads to an unbalanced dataset where the number of instances varies across different classes. Particularly, in the publicly available dataset under this study, a small portion of data points corresponds to the eating event while the majority of activities are non-eating. If trained on this skewed distribution, the classifier may learn to predict all the activities as non-eating and achieve a high accuracy level because a majority of the instances are non-eating. To handle the skewed nature of the dataset during training, we use an up-sampling technique. Specifically, in each iteration of active learning, after choosing the most informative instances, we up-sample the minority class among those selected instances by synthesizing new samples and adding a balanced set of instances to the labeled pool. For generating the synthesized instances, our offline PALS uses Synthetic Minority Over-sampling Technique (SMOTE) \cite{chawla2002smote}. Particularly, for each example of minority class, SMOTE introduces synthetic examples along the line segments of the $k$ minority class nearest neighbors.

\subsection{Performance Metrics}
As discussed previously, because of the skewed nature of the dataset, a naive classifier tends to classify all instances as the majority class (i.e., non-eating), which is usually a less important class and achieves a high accuracy. On the other hand, we cannot use the up-sampling technique for the test dataset because the test data should be a representation of the real data and needs to remain unmodified. Therefore, to avoid the disadvantage of reporting accuracy of a naive classifier, we need to consider different performance metrics than the accuracy to effectively evaluate the performance of the classifier \cite{jeni2013facing}. For this specific problem, we aim at detecting `eating moment' as the event of interest. Since the event of interest (i.e., class=`1') is the activity with minority of instances, traditional classifiers tend to have a poor recall by ignoring these important instances and predict almost everything as non-eating (i.e., class=`0'). Therefore, the binary \textit{Recall}, defined below, is an important metric in evaluation of the trained models.

\begin{equation}
Recall = \frac{True Positive}{True Positive + False Negative}
\end{equation}

However, we note that relying on \textit{Recall} alone is not enough for comparing the performance of the learned models. In particular, one can train a poor classifier by only optimizing the Recall value by predicting all instances as `eating moment'. Therefore, while optimizing the Recall, we should ensure that \textit{Precision} of the model also remains acceptable. Precision of the model is defined as follows.

\begin{equation}
Precision = \frac{True Positive}{True Positive + False Positive}.
\end{equation}

In this paper, we compute the \textit{f-score} value, which is a metric to measure the quality of the model based on the balance between \textit{Precision} and \textit{Recall}. The \textit{f-score} value is traditionally defined as   

\begin{equation}
F-score = 2 . \frac{Precision . Recall}{Precision + Recall}	
\end{equation}

\section{Results}
This section presents experimental results for offline PALS on both SW6S dataset (described in \secref{dataDesc}) and SW3S dataset (described in \secref{DColl}) as well as for real-time PALS on SW3U (described in \secref{DColl}). 

\subsection{Algorithm Design Choices}
Before evaluating the performance of our algorithms, we discuss algorithm design choices and trade-offs. Particularly, because we use a proximity graph to construct the classification model, here we discuss our approach for similarity assessment (i.e., similarity kernel). We also discuss the effectiveness of using entropy of classification as a metric for selecting instances to query during active learning. 

\subsubsection{Similarity Kernel}
As described in \secref{prox-model}, we use $k$-NN schema for constructing our similarity graph. Since the choice of similarity kernel affects the performance of the model, in this section, we examine our choice of kernel. Particularly, we compare our $k$-NN schema with the Radial Basis Function (RBF) (i.e., Gaussian) kernel \cite{nasrabadi2007pattern}. RBF kernels are popular in the field of semi-supervised learning and the similarity function ($\eta$) is defined by:
\begin{equation}
\eta(i,j) = e^{-\dfrac{\| \bf{x_i} - \bf{x_j} \|^2}{2\sigma^2}}
\end{equation}
\noindent where $\sigma$ is a free parameter that determines the {\it width} of the Gaussian kernel. 

We conducted an experiment comparing the performance of the PALS while using any of these two widely used similarity kernels. 

\begin{figure}[tbh!]
	\includegraphics[width=\linewidth]{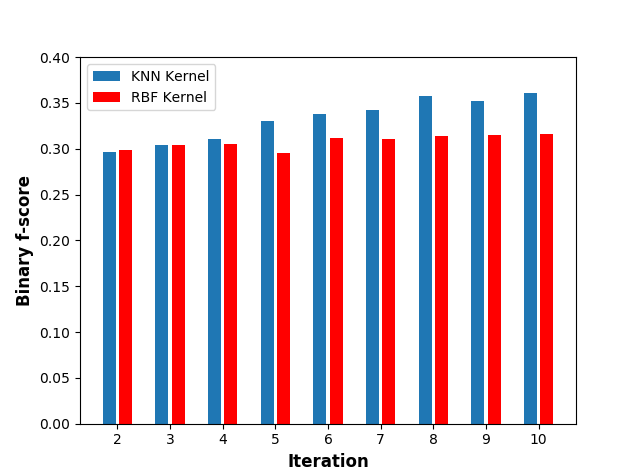}
	\caption{Performance of RBF kernel vs. $k$-NN kernel on SW3S dataset.}
	\label{fig:rvk3}
\end{figure}

\begin{figure}[tbh!]
	\includegraphics[width=\linewidth]{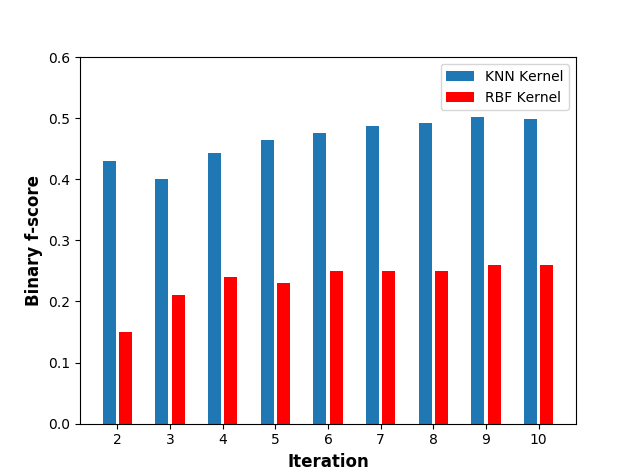}
	\caption{Performance of RBF kernel vs. $k$-NN kernel on SW6S dataset.}
	\label{fig:rvk6}
\end{figure}

As \figref{rvk3} and \figref{rvk6} illustrate, the $k$-NN kernel outperforms the RBF kernel in our application and achieves a higher f-score in detecting eating moments in all learning iterations. In particular, on SW3S dataset, models with both kernels start with a binary f-score of around $0.30$. However, the model with $k$-NN kernels reaches an f-score of more than $0.35$ only after $10$ iterations while the model with RBF kernel does not improve with more iterations. On the SW6S dataset, the model trained by $k$-NN kernel significantly outperforms the model trained using RBF kernel, as it starts with a $0.41$ f-score and reaches to $0.49$ f-score after only 10 iterations. However, the best f-score achieved by the model with RBF kernel is only $0.25$ on the SW6S dataset.

\subsubsection{Evaluation of Instance Selection}
\begin{figure}[H]
	\includegraphics[width=\linewidth]{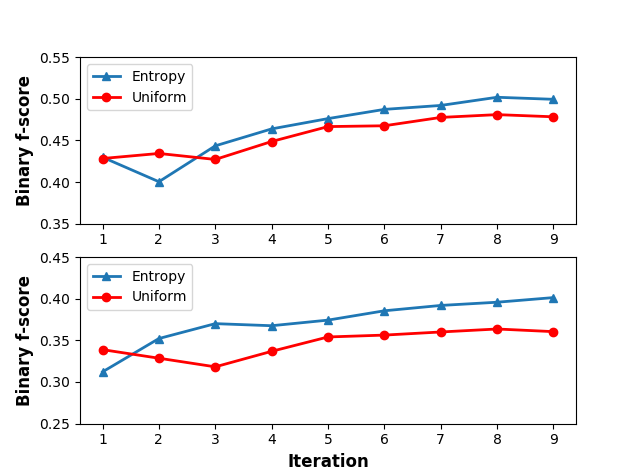}
	\caption{Entropy-based vs. uniform sampling on SW3S dataset (top) and SW6S dataset (bottom).}
	\label{fig:evr}
\end{figure}

One of the important settings in PALS implementation is that we use the entropy of classification on unlabeled instances to quantify the informativeness of an instance. Here, we show the effectiveness of our instance selection approach. We conducted an experiment comparing our instance selection approach with an approach that chooses the instances uniformly. As \figref{evr} shows, using entropy as the criteria for choosing instances results in the classifier achieving a higher f-score. On the other hand, by sampling instances from a uniform distribution, the f-score does not improve beyond few iterations. One explanation for this observation is the skewed nature of the dataset, which using uniform distribution may result in sampling more non-eating instances to be used for our active learning approach.

\subsection{Performance of Offline PALS}
In this section, we present comparison of our algorithm with prior research in the area of eating moment detection as well as state-of-the-art machine learning methods. Prior to presenting the results, we describe our comparative evaluation approach. 

Research in the area of eating gesture detection using wrist-worn inertial sensor is new. To the best of our knowledge, the most recent successful approach presented by E. Thomaz and et. al. \cite{thomaz2015practical} which tackles the similar problem of food intake gesture recognition as this paper. The classifer built in  \cite{thomaz2015practical} uses Random Forest algorithms with the following settings. They used Scikit-learn Python package\cite{pedregosa2011scikit} implementation of Random Forest with number of trees in the forest set to 185. For the rest of this paper we call their approach RFA.

Additionally, we compare our algorithm to a classifer built using XGBoost learning algorithm\cite{chen2016xgboost}. XGBoost has recently been dominating the field of applied machine learning and used to win the Kaggle \footnote{'https://www.kaggle.com'} competitions in recent years. Furthermore, XGBoost was used in all top-10 winning teams in KDDCup 2015\cite{chen2016xgboost}. XGBoost is an optimized and distributed implementation of Gradient Boosting. It provides a parallel tree boosting method to effectively solve machine learning problems in the industrial scale. For this experiment, we used the open source implementation of XGBoost.

To conduct the offline comparison, we suppose that each algorithm have access to the $20$\% of in-lab data as its training set and we use the remaining $80$\% of the data as test set to validate the performance of the algorithm. As shown in \tblref{palvsothers}, offline PALS outperforms both other approaches in correctly classifying eating moments. Specifically, offline PALs can achieve to 41\% and 48\% f-score when running on SW3S and SW6S datasets, respectively, which is a good improvement over XGBoost and RFA.
Also, low recall for eating class refers to the classifier having a high bias in classifying all instances as not-eating. This again emphasizes the importance of selecting appropriate metrics while working with skewed datasets. As presented in \tblref{palvsothers}, offline PALS achieves a $62$\% and $64$\% recall when running on SW3S and SW6S datasets, respectively. These numbers demonstrate significant improvements over RFA and XGBoost classifier.

\begin{table}[tbh!]
\centering
\caption{Performance of offline PALS vs. other approaches.}
\begin{tabular}{l{l}{c}{c}}\hline \hline
 & &  recall & binary f-score \\ \hline \hline \\
   & Offline  PALS &  0.62 &  0.41   \\ 
 
 SW3S dataset &XGBOOST &  0.25 &  0.35  \\ 

 &RFA &  0.22 &  0.34  \\  
 \hline \\
&Offline  PALS &  0.64 &  0.48   \\ 
SW6S dataset &XGBOOST  &  0.32 &  0.40  \\ 

&RFA  & 0.10 &  0.18  \\   \hline
\end{tabular}
\label{tbl:palvsothers}
\end{table}

\subsection{Performance of Real-Time PALS}
To the best of our knowledge, there is no prior algorithm for real-time training of eating-moment recognition in real-life settings. Therefore, for the purpose of evaluation, we conducted two experiments highlighting the effect of query budget and decision threshold estimation on the performance of the real-time PALS algorithm.

\subsubsection{Query Budget}

\begin{figure}[tbh!]
  \includegraphics[width=\linewidth]{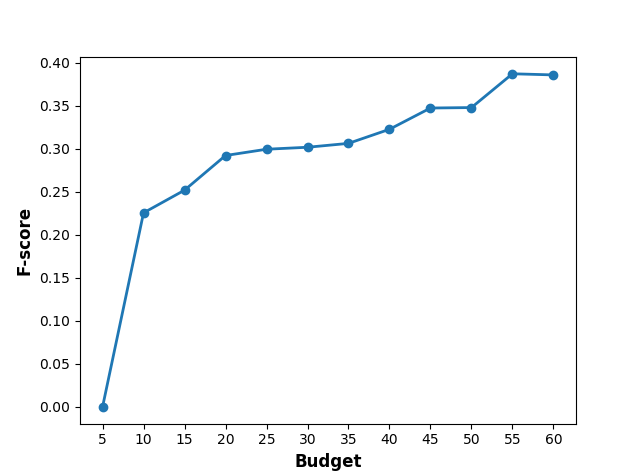}
  \caption{Performance of the learned model in terms of f-score as a function of query budget on SW3U dataset.}
  \label{fig:QBE}
\end{figure}

In this experiment, we assess the effect of query budget, $\Delta$, on the performance of real-time PALS approach in classifying eating moments. To this end, we examined twelve different values of query budget per hour for different subjects in in-the-wild setting on SW3U dataset. \figref{QBE} shows the f-score value averaged over all the subjects at the end of training cycles.

As \figref{QBE} illustrates, increasing the value of the budget helps in obtaining a more personalized classifier for each participant and leads to a higher performance measure. In particular, the average f-score starts at around $0$ with $\Delta=5$ queries per hour and reaches a value of $39$\% when the budget has increased to 60 queries per hour.

With very limited query budget to query the user in real-time, the model cannot adapt itself from the lab-setting to real-world setting and the f-score of eating moment is less than $1$\%.  In other words, only relying on the model trained on in-lab collected data, the model tends to detect all activities as non-eating. One reason behind this is that the distribution of eating vs. non-eating activities is very different from lab setting to real-world setting. Also, it means that in real-world setting and without any constraints, people tend to perform eating activity very different than how they are doing in the lab settings. This result again highlights the importance of designing the adaptive models for real-world settings. 

Furthermore, only considering a small query budget, the model gets a significant gain. Particularly, increasing the query budget to $10$ queries, the average f-score of detecting eating-moments increases by around $23.1$\%. We see the constant improvement of model performance by increasing the query budget. Particularly, on average over all subjects, the model reaches the f-score of $29.8$\% when the maximum query budget set to the $20$ queries. While still improving, the rate of performance improvement decreases for query budgets more than $20$ and the model achieves the f-score of $39$\% by having the query budget of $60$.  

There is always a trade-off between the query budget and user convenience. While by increasing the query budget, we increase the performance of the model, we may also increase the risk of user inconvenience. 

\subsubsection{Comparison of Thresholding Methods}
As described in \secref{ATS}, decision threshold is used by the classifier to determine if the current instance of activity is valuable enough to query the user and obtain its label. To verify the effectiveness of our approach in updating the decision threshold, $\lambda$, we designed two different methods for governing the value of decision threshold. In the first method, the value of the $\lambda$ is learned from the in-lab training data which is derived based on the ratio of budget to the size of the dataset. Since, this value extracted from the in-lab data and remains unchanged during the real-time training, we refer to the decision threshold obtained in this approach as {\it static $\lambda$}. The second method uses the knowledge of best possible value for the threshold in a time interval to select the most informative instances based on the entropy of the classification decision. This experiment provides an experimental upper-bound for the adaptive lambda because it has unlimited access to the future data and can extract the most accurate value of $\lambda$ that the adaptive lambda algorithm attempts to estimate. We refer to the decision threshold obtained by this approach as {\it best $\lambda$}.
 
\begin{figure}[tbh!]
  \includegraphics[width=\linewidth]{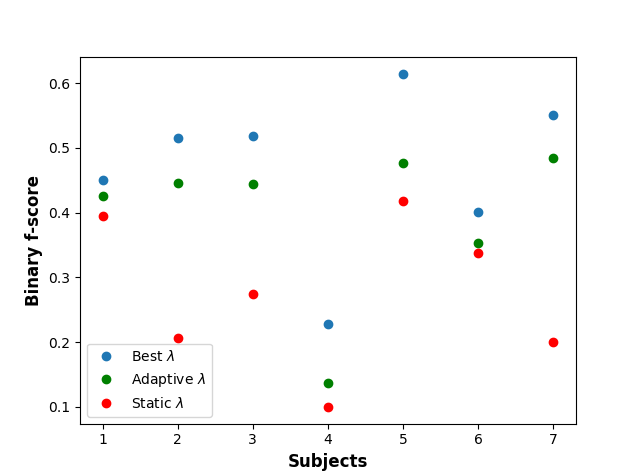}
  \caption{Comparison of {\it best $\lambda$}, {\it adaptive $\lambda$}, and {\it static $\lambda$} approaches for decision threshold in term of f-score on SW3U dataset.}
  \label{fig:ALE}
\end{figure}

In this experiment, we compared the performance of the eating moment detection models trained on real-time data of SW3U dataset using {\it best $\lambda$}, {\it adaptive $\lambda$}, and {\it static $\lambda$}. The x-axis refers to different subjects and the y-axis shows the binary f-score value for classifying eating class instances. The query budget is set to $60$ queries per hour for this experiment. 
As \figref{ALE} shows, adaptive lambda algorithm achieves performance values close to the {\it best $\lambda$} while using a static value for $\lambda$ performs poorly across different subjects. Specifically, {\it adaptive $\lambda$} on average can achieve to 7\% less f-score compared to {\it best $\lambda$} and 12\% better f-score compared to {\it static $\lambda$}. Also, to evaluate the extreme cases, {\it adaptive $\lambda$} achieves to 13\% less f-score compared to {\it best $\lambda$} for subject number 5 while it works better for other subjects. Furthermore, {\it adaptive $\lambda$} works, in worst case, slightly better than {\it static $\lambda$} with 1.6\% better f-score for subject number 6 while it outperforms {\it static $\lambda$} for other subjects specifically subject 7 with 28\% higher f-score. To summarize the results of this experiment, {\it best $\lambda$}, {\it adaptive $\lambda$}, and {\it static $\lambda$} on average can provide 47\%, 39\%, and 28\% average f-score for all subjects of SW3U dataset.

\section{Discussion and Future Work}
We developed a novel proximity-based approach for recognizing eating gestures with the goal of significantly decreasing the need for obtaining labeled data from the users. One challenge of the proposed approach is that it assumes that the pattern of the minority class (i.e., eating) remains unchanged over time. Therefore, if the user's activity pattern changes due to changes in life style or user being interested in a new type of food, the method may not easily detect the new eating patterns. On the other hand, continuous learning with the same rate might not be feasible since the user needs a stable model after the training phase. Facilitating a trade-off between exploitation of the learned model and exploration of new activity patterns is an interesting future research direction. A potential approach to address this problem is to examine how one can leverage reinforcement learning paradigm to handle the exploration/exploitation trade-off.

In this study, we used an off-the-shelf smartwatch to detect  `eating moment' activities. Our future research also involves studying the utility of other wearable and non-wearable sensory devices for eating moment detection through active learning.

While the focus of this study was on eating moment detection, we expect that the methodologies developed in this project can be applied to a broader class of activity recognition applications. In the future, we plan to study the effectiveness of PALS in devising personalized activity recognition algorithms. This way, one can integrate diet monitoring capabilities with the ability to recognized daily activities and develop a smart-health coach.

\section*{Acknowledgment}
This work was supported in part by the United States National Science Foundation, under grant CNS-1932346. Any opinions, findings, conclusions, or recommendations expressed in this material are those of the authors and do not necessarily reflect the views of the funding organizations.

\section{Conclusion}
Most current approaches to detect eating moment require multiple on-body sensors or specialized devices such as neck-collars for swallow detection that are impractical for everyday usage. The goal of this research was to design a practical solution for eating moment detection. We used an off-the-shelf smartwatch that records inertial sensor data to design a non-intrusive detection system with the machine learning algorithm personalized for the end-user.

Because people perform the same activity in different manners, relying on a model that is trained on in-lab data collected of different subjects leads to a significant performance drop. In this paper, we proposed PALS ({\it Proximity-Based Active Learning on Streaming Data}), a novel proximity-based model for recognizing eating gestures. We showed that PALS significantly decreases the need for labeled data with new users leveraging active learning under limited query budget while utilizing unlabeled data. Our extensive analysis on data collected from real-subjects showed that compared to the state-of-the-art approaches, PALS,on average, achieves to 40\% higher recall and 12\% higher f-score in detecting eating events. Furthermore, we showed the effectiveness of our adaptive thresholding method and how online PALS algorithm could be adapted in the real-world settings with only limited query budget. 

\bibliographystyle{IEEEtran} 
\bibliography{refs} 

\end{document}